\documentclass[sigconf]{acmart}

\AtBeginDocument{%
  }

\setcopyright{acmlicensed}
\copyrightyear{2025}
\acmYear{2025}
\acmDOI{XXXXXXX.XXXXXXX}

\acmConference[WSDM '25]{The 18th ACM International Conference on Web Search and Data Mining}{March 10--14, 2025}{Hannover, Germany}
\acmISBN{978-1-4503-XXXX-X/18/06}

\DeclareMathOperator*{\argmax}{arg\,max}

\usepackage{algorithm}
\usepackage{algorithmic}
\usepackage{makecell}

\usepackage{tabularx}
\usepackage{xcolor}
\usepackage{mathtools}
\DeclarePairedDelimiter\ceil{\lceil}{\rceil}

\begin{document}

\title{Reducing Hyperparameter Tuning Costs in ML, Vision and Language Model Training Pipelines via Memoization-Awareness}



\author{Abdelmajid Essofi$^{ 1\ast}$, Ridwan Salahuddeen$^{1\ast}$, Munachiso Nwadike$^1$, Elnura Zhalieva$^1$, Kun Zhang$^1$, Eric Xing$^1$, Willie Neiswanger$^2$, Qirong Ho$^1$}
\affiliation{
\institution{$^1$Mohamed Bin Zayed University of Artificial Intelligence ~~~~~$^2$University of Southern California}
\country{}
}
\affiliation{
\institution{$^1$\{adbelmajid.essofi, ridwan.salahuddeen, munachiso.nwadike, elnura.zhalieva, kun.zhang, eric.xing, qirong.ho\}@mbzuai.ac.ae~~~~~ $^2$neiswang@usc.edu}
\country{}
}
\authornote{Both authors contributed equally to this research.}

\renewcommand{\shortauthors}{Essofi et al.}

\begin{abstract}


The training or fine-tuning of machine learning, vision and language models is often implemented as a pipeline: a sequence of stages encompassing data preparation, model training and evaluation.
In this paper, we exploit pipeline structures to reduce the cost of hyperparameter tuning for model training/fine-tuning, which is particularly valuable for language models given their high costs in GPU-days. We propose a "memoization-aware" Bayesian Optimization (BO) algorithm, EEIPU, that works in tandem with a pipeline caching system, allowing it to evaluate significantly more hyperparameter candidates per GPU-day than other tuning algorithms. The result is better-quality hyperparameters in the same amount of search time, or equivalently, reduced search time to reach the same hyperparameter quality.
In our benchmarks on machine learning (model ensembles), vision (convolutional architecture) and language (T5 architecture) pipelines, we compare EEIPU against recent BO algorithms:
EEIPU produces an average of
$103\%$ more hyperparameter candidates (within the same budget), and increases the validation metric by an average of $108\%$ more than other algorithms (where the increase is measured starting from the end of warm-up iterations). 
\end{abstract}





\settopmatter{printacmref=false}
\setcopyright{none}
\renewcommand\footnotetextcopyrightpermission[1]{}
\pagestyle{plain}

\maketitle

\vspace{-2mm}
\section{Introduction}
\label{introduction}

An AI pipeline, in the context of training or fine-tuning machine learning, vision and language models, is a sequence of stages containing different types of data-consuming or data-processing code.
Examples of AI pipeline stages include data collection and preprocessing, model training/fine-tuning (possibly involving several models), distillation, validation and deployment. For instance, the T5 language model~\cite{2020t5} distillation pipeline in our experiments consists of five stages: data preprocessing, teacher model fine-tuning, followed by three checkpointed stages for student model distillation.
Due to its smaller size relative to other language models, the T5 suite of text-to-text models excels in practical applications that target a well-defined and narrow application or use case, while keeping inference costs as small as possible. Because of T5's efficient trade-off between model size and fine-tuned performance, it continues to see use in natural language applications such as spam detection \cite{labonne2023spam}, text-to-sql parsing \cite{li2023graphix}, text ranking \cite{zhuang2023rankt5}, medical text processing \cite{lehman2023clinical} and more \cite{hwang2023ensemble} \cite{adewumi2023t5}. 

Hyperparameter tuning algorithms are useful automatic tools for improving model quality; however, such algorithms require at least dozens of pipeline runs to yield models with superior quality versus human-chosen hyperparameters. Consider the smaller models from the T5 family, ranging from 60m to 770m parameters: a single fine-tuning run takes between several hours to days on one GPU, depending on the dataset and model size. Thus, hyperparameter tuning would require days to weeks, which negates the benefit of automation. The problem is even more pronounced with larger language models; for example, training 7B models requires $80$\textit{k} to $130$\textit{k} GPU hours \citep{liu2023llm360, touvron2023llama}, with an an estimated dollar cost of \$$410$\textit{k} to \$$688$\textit{k}. This motivates our goal of reducing hyperparameter tuning costs in AI pipelines.

Because pipelines are sequential -- each stage's output is the input of the next one -- we can memoize (cache) the outputs of intermediate stages. This enables the development of ``memoization-aware" hyperparameter tuning algorithms, which reduce search costs by re-starting in the middle of the pipeline, rather than the beginning.
We pick Bayesian Optimization (BO) ~\citep{snoek2012practical, lee2020cost, abdolshah2019cost, multi-step-budg-bo, BAPI, lin2021bayesian, kusakawa2021bayesian} as a starting point to develop our memoization-aware method, for two reasons: (1) many AI pipelines contain non-differentiable code (e.g. data preprocessing or model output postprocessing) or consist of multiple models (e.g. model ensembles, student-teacher model distillation), which precludes the use of gradient-based tuning methods~\citep{maclaurin2015gradientbasedhyperparameteroptimizationreversible, micaelli2021gradientbasedhyperparameteroptimizationlong, pmlr-v70-franceschi17a, pmlr-v48-luketina16}; (2) the cost (time to run) a hyperparameter configuration needs to be modeled, because it is (a) unknown beforehand, and (b) can change with hyperparameter choices (e.g. data augmentation, number of fine-tuning epochs) -- as we shall see, BO can be extended to model such unknown and changing costs.



BO uses previously collected observations $\{x_i, y_i\}_1^N$ (in our context, $x_i$ are hyperparameter configurations, and $y_i$ is a measure of model/pipeline quality, like accuracy) to train a surrogate model that approximates the function landscape, guiding the search for optimal solutions through a series of evaluations $f(x_1), f(x_2), ..., f(x_N)$. A key strength of BO lies in its ability to balance the exploration of new areas of the search space with the exploitation of known promising regions, known as the exploration-exploitation trade-off. However, many BO hyperparameter tuning algorithms assume {\it uniform costs} for each evaluation (hyperparameter configuration); as we remarked earlier, this is not the case for some AI pipelines.
Consider a hyperparameter that controls the number of fine-tuning epochs: a naive BO tuning algorithm will repeatedly increase the number of epochs, because it yields increasing model quality (albeit with diminishing returns). Unfortunately, this obviously causes pipeline runs to take increasingly longer (higher cost), resulting in two problems: (1) the search budget becomes exhausted early, with fewer hyperparameter configurations evaluated; (2) when there are many different hyperparameters, the algorithm may fail to explore some of them, due to running out of search budget.

\begin{figure}[t]
\centering
 \includegraphics[width=\linewidth]{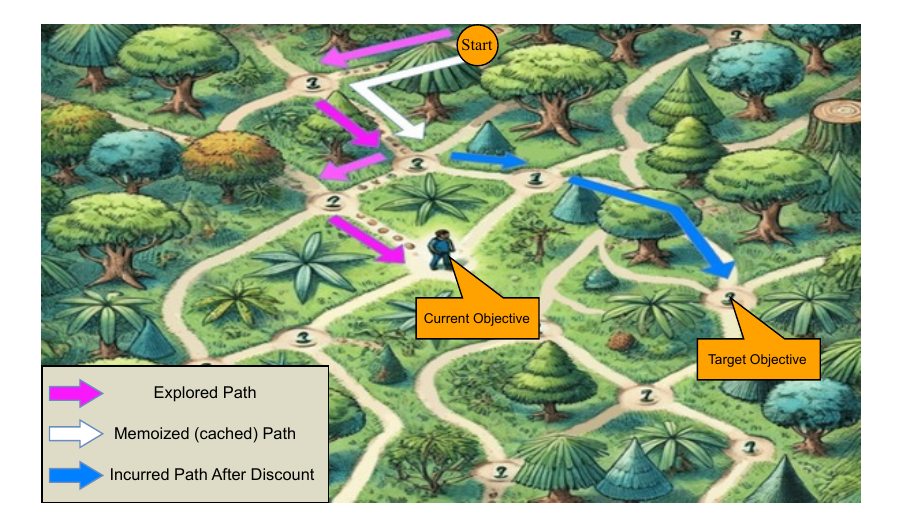}
 \caption{Memoization-aware hyperparameter tuning, explained via a ``breadcrumb" analogy. By caching pipeline stage outputs from earlier hyperparameter runs (the breadcrumbs), the cost of hyperparameter search on later stages is reduced {\it if they reuse the earlier stage outputs}. Our goal is to reduce the high cost of hyperparameter search in language, vision and ML pipelines; memoization-aware algorithms achieve this by reusing cached stage outputs, exploring later-stage hyperparameters at a fraction of their regular cost.
 }
\label{fig:forest}
\end{figure}

These challenges with naive BO can be overcome by introducing {\it memoization-awareness}. Consider the analogy in Figure~\ref{fig:forest}, portraying hyperparameter search as a journey through a dense forest, where each path could either lead to discovery or delay, and different paths incur different time costs. To exit the forest, we need clever
navigational techniques, such as leaving breadcrumbs to retrace our steps quickly.
Memoization-awareness is to pipeline hyperparameter search what breadcrumbs are to forest explorers.


In order to make BO memoization-aware, we must incorporate evaluation costs into the candidate evaluation process, making it {\it cost-aware}. All BO hyperparameter tuning algorithms construct a surrogate model to predict model quality, typically a Gaussian Process (GP) \citep{gp}. This model is used as a component of an {\it acquisition function}, which the BO algorithm searches (maximizes) over to pick the next hyperparameter configuration for evaluation. A common acquisition function is Expected Improvement (EI) \citep{ei,sim},
defined as the average (over all functions modeled by the GP) improvement of a potential new hyperparameter configuration $x$ over the best one seen so far $x^*$, i.e. $\text{EI}(x) = \mathbb{E}[\max(f(x) - f(x^*), 0)]$.
EI does not model or make use of evaluation costs in any form, and previous works have shown that, unsurprisingly,
EI is suboptimal when hyperparameter tuning budgets are expressed as a total cost over all evaluations, rather than a fixed number of evaluations.
The challenge with modeling evaluation costs, is that said costs are non-uniform (e.g. a hyperparameter controlling epoch count) and initially unknown; hence, they need to be modeled using only evaluations seen so far. Early cost-aware approaches, such as the Expected Improvement Per Second (EIPS) \citep{snoek2012practical} and Cost Apportioned BO (CArBO) \citep{lee2020cost}, approach this by fitting a second GP to model log-costs $\ln c(x)$ (the log ensures costs are always positive).

\begin{figure*}[ht]
\centering
 \includegraphics[width=0.95\linewidth]{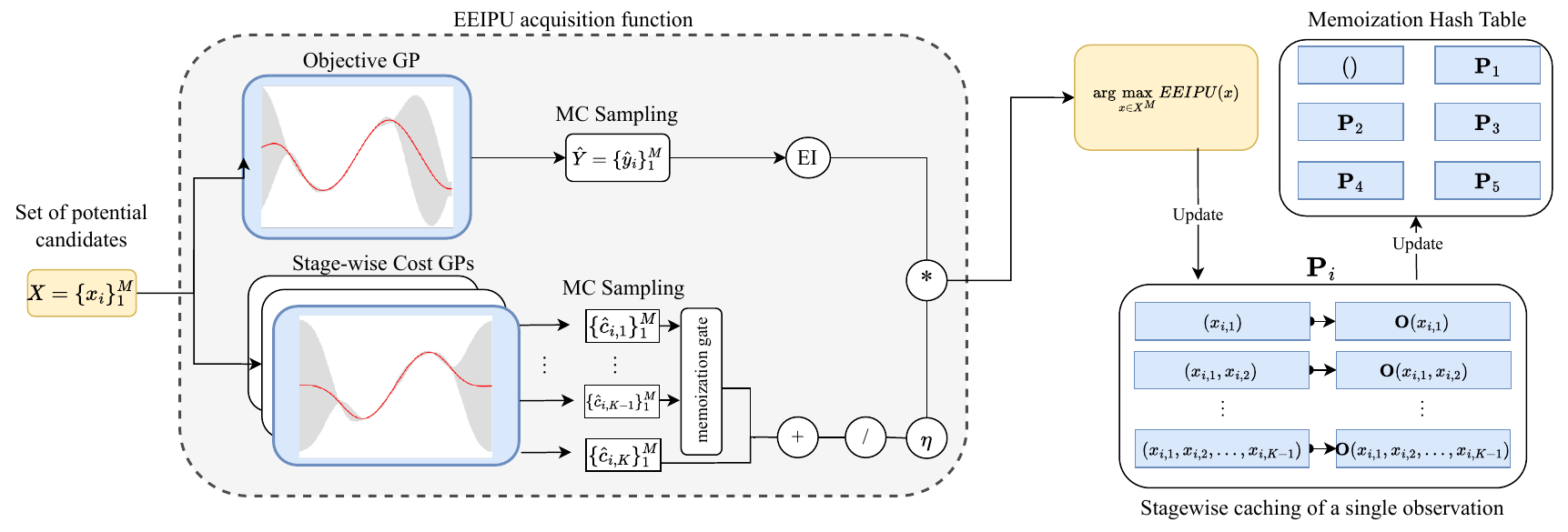}
 \vspace{-2mm}
 \caption{Our EEIPU algorithm plus memoization (caching). Our method fits $k+1$ GP models (one objective model and $k$ stage-wise cost models), given: (1) a set of $N$ observations (previous hyperparameter runs), (2) their corresponding objective values (model quality/validation scores) and observed runtime costs (on each of the $k$ pipeline stages), and (3) the remaining optimization budget. Each iteration of EEIPU selects a new hyperparameter candidate as follows: first, we randomly generate $M$ potential hyperparameter candidates $X^M$, preferring to re-use hyperparameter prefixes currently in the memoization cache. Then, we estimate these $M$ candidates' objective and stage-wise costs through MC sampling on the respective GP models.
 Memoized stages, if any, have their costs discarded by the \emph{memoization gate}, leaving only non-memoized stages when compting the expected inverse cost $\mathbb{E}[1/C(x)]$. This expected inverse cost is raised to the power $\eta$ as a ``cost-cooling" mechanism~\citep{lee2020cost}, which shifts attention towards high-objective exploitation over low-cost exploration as the search budget depletes. Multiplying (1) with (2) gives the EEIPU acquisition function, which we maximize (by simply ranking all $M$ candidates) to return the best hyperparameter candidate for evaluation. Finally, the cache is updated to hold the ``hyperparameter prefix set" of the top-$5$ performing candidates seen so far, along with their stage-wise outputs $\mathbf{O}(x_{i,k}) \forall k \in \{1,2,...,K-1\}$. See Section~\ref{sec:methods} for details.}
\label{fig:eeipu}
\end{figure*}

To develop a BO hyperparameter tuning method that is (1) cost-aware and (2) employs memoization to reduce search costs, we propose the Expected-Expected Improvement Per Unit-cost (EEIPU) acquisition function (Figure~\ref{fig:eeipu}). EEIPU extends EI, making it cost aware by incorporating stage-wise cost models,
and memoization aware by using a cache to store pipeline stage outputs from previous hyperparameter evaluations.
When modeling the costs of new hyperparameter configurations $x$, the configurations whose {\it hyperparameter prefixes} match the cached stages have their predicted costs discounted from the evaluation process. Accordingly, we refer to the cached stages as a ``prefix set" $\mathbf{P}$, where a cached stage at position $k$ in the pipeline is the result of a hyperparameter prefix $x_1,\dots, x_k$ over the first $k$ stages.
By proposing new configurations $x$ that prefix-match some stage in the cache, EEIPU can explore the hyperparameter space ``around" said stage at a fraction of the original evaluation cost.
When evaluated against recent BO methods, EEIPU outperforms them on both total evaluations and achieved model quality, achieving 103\% more evaluations on real pipelines (148\% on synthetic) within the given time budget, and 108\% higher model quality after warm-up iterations (58\% on synthetic).
Our key contributions are summarized:
\begin{itemize}
\item We propose EEIPU, a cost- and memoization-aware acquisition function that adapts EI to the multi-stage pipeline setting. EEIPU fits a separate log-cost GP for every stage of the pipeline. These GPs are used to calculate the expected total inverse cost, where the total cost $\hat{C}(x)$ is the sum of individual stage costs in the pipeline. Thus, $\text{EEIPU}(x)\triangleq \text{EI}(x)\times\mathbb{E}[\frac{1}{\hat{C}(x)}]^{\eta}$.
    
\item EEIPU employs a cost-cooling factor $\eta$ that decreases with the remaining search budget.
This encourages EEIPU to explore many low-cost regions initially, before switching over to exploiting high-quality regions (that may have high costs).
    
\item EEIPU's cache retains only the stage outputs from the top-$Q$ highest-performing hyperparameter configurations $x$. This reduces the overhead time required for EEIPU computations, and encourages EEIPU to explore high-performing, low-cost regions of the space around the top-$Q$ configurations. In our benchmarks, $Q=5$ produced the best results.

\end{itemize}
We provide implementation details in this repository \url{https://github.com/anonymousWSDMSubmission/cost-bo}.
\vspace{-1mm}

\section{Background and Related Work}
\label{background}

Pipeline execution systems are used to optimize ML workflows or AI pipelines. \citet{10.14778/3342263.3342633} introduce an intermediate representation in Lara, optimizing both data preprocessing and ML computations through unified operations on relational and linear algebra. \citet{267386} focus on Spark's Resilient Distributed Datasets (RDDs) to manage data effectively across iterative ML processes, highlighting the benefits of optimized data handling for repetitive computational tasks.
EEIPU can be implemented on top of any pipeline system that supports caching, and is orthogonal to such work.


An early work addressing cost-aware BO is \citet{snoek2012practical}, who introduce the expected improvement per second (EIPS) acquisition function.
\citet{lee2020cost} improved on this method in their work on Cost-Aware Bayesian Optimization (CArBO), where they observed a bias of EIPS towards inexpensive points.
Both works handle BO in the unknown-cost setting by fitting a single GP for modeling costs, and cannot support memoization-awareness which requires each pipeline stage's cost to be modeled. In contrast, EEIPU models every pipeline stage's cost with a different GP.

Other BO approaches that handle unknown costs include \citet{multi-step-budg-bo} and \citet{BAPI}
which use non-myopic policies to choose the next evaluation point.
However, it is not obvious how memoization-awareness can be incorporated into these non-myopic policies. \citet{abdolshah2019cost} introduce cost-aware multi-objective BO (CA-MOBO) in a single-stage setting; their method handles non-uniform (but known) cost landscapes by specifying cost-aware constraints over the search space.
Because CA-MOBO requires prior knowledge about the cost landscape, it is inapplicable to our AI pipeline setting.

Two related works \citep{lin2021bayesian,kusakawa2021bayesian} address the multi-stage pipeline setting, but have shortcomings. \citet{lin2021bayesian} employs a slowly moving bandit (SMB) algorithm that uses a tree structure to model the hyperparameter space.
Their approach, LaMBO, either requires users to specify partitions within the hyperparameter space, or else it defaults to a bisection heuristic that unfortunately causes the tree to grow exponentially with more hyperparameters. Furthermore, each LaMBO iteration requires solving a local BO problem for every leaf in the tree. These factors make LaMBO inefficient for longer pipelines with many hyperparameters.
\citet{kusakawa2021bayesian} consider a multi-stage BO (MS\_BO) process, where every stage has an objective function modeled by an independent GP.
However, MS\_BO does not consider evaluation costs (i.e. it is not cost-aware). Furthermore, it requires that every stage to output its own objective value; this makes it unsuited for AI pipelines where the objective (e.g. model quality) is known only after the final stage.

Table \ref{table:ablation} summarizes our contributions and key differences between our method and related works.

\begin{table}[h]
\centering
\caption{EEIPU and comparable BO methods}
\begin{small}
\begin{tabular}{lccccr}
\toprule
\makecell{Acquisition \\ function} & \makecell{Cost- \\ aware} & \makecell{Unknown \\  costs} & \makecell{Multi- \\ stage} & \makecell{Memoization- \\ aware}  \\
 \midrule
 EI \citep{ei} & No & No & No & No \\
 CArBO \citep{lee2020cost} & \textbf{Yes} & \textbf{Yes} & No & No \\
 EIPS \citep{snoek2012practical} & \textbf{Yes} & \textbf{Yes} & No & No \\ 
 MS\_BO \citep{kusakawa2021bayesian} & No & No & \textbf{Yes} & \textbf{Yes} \\  
 LaMBO \citep{lin2021bayesian} & No & No & \textbf{Yes} & \textbf{Yes} \\
 \textbf{EEIPU} & \textbf{Yes} & \textbf{Yes} & \textbf{Yes} & \textbf{Yes} \\ 
 \bottomrule
\end{tabular}
\end{small}
\label{table:ablation}
\end{table}
\section{Methodology}
\label{sec:methods}

We consider a "black-box cost" multi-stage AI pipeline setting, where pipeline stages are run sequentially before producing a final output. Our goal is to maximize an unknown objective function $f$ with no analytical form or gradient information. Let $\mathcal{X}_k$ denote the variable space of the $k^{th}$ stage in a $K$-stage pipeline, and $\mathcal{X} = \mathcal{X}_1 \times \mathcal{X}_2 \times ... \times  \mathcal{X}_K$ denote the variable space across all stages of the pipeline. Accordingly, we want to find the optimizer
\[x^* \in \argmax_{x \in \mathcal{X}} f(x).\]
When a hyperparameter configuration $x \in \mathcal{X}$ is executed through an AI pipeline, we obtain a noisy output $y = f(x) + \epsilon$ as an objective value and incur a total cost $C = C(x) + \epsilon$, measured by the wall-clock time of running the hyperparameter combination through the pipeline. Bayesian Optimization (BO) is a sample-efficient technique that uses a set of candidate observations $X = \{x_1, x_2, \dots, x_N\}$ along with their observed values $Y = \{y_1, y_2, \dots, y_N\}$ to build a surrogate model of the unknown function in search of the optimizer $x^*$. 

EEIPU evaluates a candidate hyperparameter combination $x$ by scaling its expected improvement EI$(x)$ by its corresponding expected cost $\hat{C}(x)$. Thanks to our memoization-aware setting,
EEIPU can choose to resume from a cached intermediate stage, whilst using new hyperparameters for the remaining stages. This allows for more observations $x\in X$, because EEIPU does not always have to incur the full pipeline cost.

We incorporate memoization-awareness into EEIPU by training a set of $K$ GP models on $(X_k, \ln(C_k)) \; \forall k \in \{1, 2, ..., K\}$ for modeling stage-wise cost surrogates, where $X_k$ is the set of hyperparameters run through the $k^{th}$ stage of the pipeline, $C_k$ is the incurred cost, and the natural log is applied to enforce positivity on the GP model predictions. Note that these $K$ GPs are in addition to the usual BO GP on $(X, Y)$ for modeling the objective function $f$. We choose to model the different pipeline stage costs as independent of each other; accurate discovery of such cost dependencies is impractical given the limited search budgets in hyperparameter tuning.

The aforementioned $K$ GPs
are used to predict the costs of each non-memoized stage when computing EEIPU$(x)$. Next, we explain in detail EEIPU's memoization- and cost-awareness (Figure~\ref{fig:eeipu}).

\begin{algorithm}[h]
    \caption{Expected-Expected Improvement Per Unit}
    \label{alg:eeipu}
    \begin{algorithmic}
        \STATE {\bfseries Input:} Prefix set $\mathbf{P}$, objective GP$_y$, cost GP$_{c_k} \forall k \in\{1,2,...,K\}$, num. of candidates $M$, num. of MC samples $D$, total budget \texttt{tot\_bgt}, consumed budget \texttt{used\_bgt}, hyperparam. generation bounds \texttt{hp\_bounds}
        \STATE $N = \text{len}(\mathbf{P})$, \texttt{b\_size} $= M/N$
        \FOR{\texttt{pref} in $\mathbf{P}$}
        \STATE $\delta \gets$ \text{len}(\texttt{pref}), \quad\quad\quad\quad\quad\quad\quad\quad\quad\;\;\textcolor{blue}{num. of cached stages}
        \STATE $\texttt{cand} \sim Unif(\texttt{hp\_bounds}, M)$ \quad\quad\;\; \textcolor{blue}{generating candidates}
        \STATE $\texttt{cand}[:,:\delta] = \texttt{pref}$ \;\;\;\;\;\;\;\;\;\;\; \textcolor{blue}{retain memoized prefixes}
        \FOR{$i$ in range(\texttt{b\_size})}
        \STATE $x \gets \texttt{cand}[i]$, $\hat{y} \gets \text{GP}_y(x, D)$
        \STATE \textbf{for} $s$ in range($\delta$) \textbf{ do } $\hat{c}_s \gets \epsilon$
        \STATE \textbf{for} $s$ in range($\delta, K$) \textbf{ do } $\hat{c}_s \gets \text{GP}_{c_s}(x[s], D)$
        \STATE Compute EI($x$) 
        \STATE Compute $\text{I}(x)$ \quad\quad\quad \textcolor{blue}{using the process described in \ref{sec:cost_aware}}
        \STATE \texttt{rem\_bgt = tot\_bgt - used\_bgt}
        \STATE $\eta \gets \frac{\texttt{rem\_bgt}}{\texttt{tot\_bgt}}$
        \STATE $\text{EEIPU}[i] \gets \text{EI} \times \text{I}^{\eta}$
        \ENDFOR
        \STATE \textbf{return} \texttt{argmax}(EEIPU)
        \ENDFOR
    \end{algorithmic}
\end{algorithm}

\subsection{Memoization-Awareness}
\label{sec:memoization}

\begin{figure}[t]
\centering
 \includegraphics[width=\linewidth]{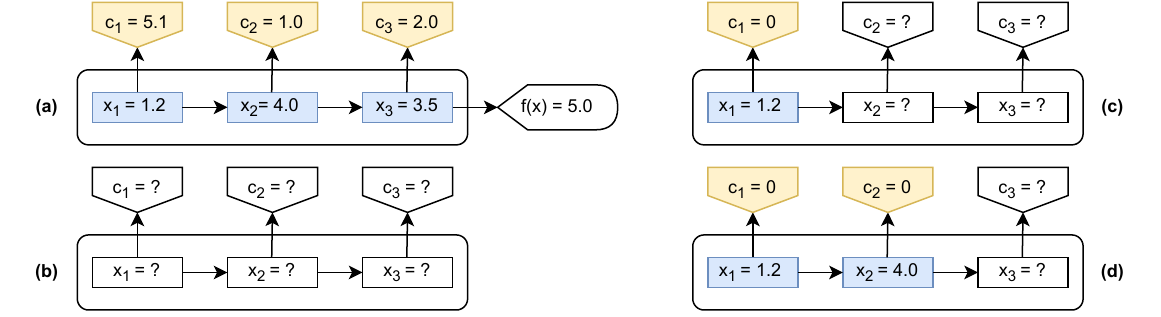}
 \caption{Prefix pooling for 1 observation (subfigure (a)) in a 3-stage pipeline. Each stage has a single parameter, respectively represented by  $x_1,x_2$, and $x_3$, with corresponding costs $c_1,c_2,c_3$. Since the query in (a) is completed, both the cost and output value of each stage are stored. Subfigures (c) \& (d) show how first 2 stages of the observation are cached as prefixes to avoid rerunning them. "?" indicates un-memoized stages, and the empty prefix (subfigure (b)) is used for open-ended search. This process is independently applied to every observation chosen by the acquisition function.}
\label{fig:mem}
\end{figure}

EEIPU's memoization approach caches the intermediate stage outputs of previously evaluated candidate points $x$, allowing EEIPU to selectively avoid rerunning earlier stages.
We begin by defining the ``prefix set" $\mathbf{P}$. Consider a set of $N$ previously evaluated hyperparameter combinations. Each combination is made up of $K$ subsets, where $x_{i,k}$ is the ensemble of hyperparameters of the $i^{th}$ observation corresponding to the $k^{th}$ stage of the pipeline. After monitoring the average number of iterations and frequency of updating the best objective value in a standard BO process, we decided to define $\mathbf{P}$ as the set of observations corresponding to the $5$-highest objectives. This value was chosen for its efficacy in enabling memoization to outperform competing methods, and the insignificance of improvement that higher numbers of cached observations bring to EEIPU's performance. For these chosen observations, we cache every one of the first $K-1$ stages, along with the output of the last-cached stage, for a total of $5 \times (K-1)$ cached prefixes. In other words, we memoize an observation $x_i$ by caching the following configurations:
$[(x_{i,1},\mathbf{O}(x_{i,1}))], [(x_{i,1}, x_{i,2}, \mathbf{O}(x_{i,1}, x_{i,2})],$ $\ldots$ $,  [(x_{i,1},$ $\ldots$ $, x_{i,K-1}, \mathbf{O}(x_{i,1},$ $\ldots$ $, x_{i,K-1}))]$, where $\mathbf{O}(x_{i,1},$ $\ldots$ $, x_{i,k})$ is the output state of stage $k$ after running the prefix $(x_{i,1},$ $\ldots$ $, x_{i,k})$ through the pipeline. This separate dataset is built and stored for every evaluated candidate point, in addition to the empty prefix $[()]$ (which ensures EEIPU will explore entirely new configurations). See Figure~\ref{fig:mem} for a 3-stage pipeline example.

When choosing a new observation, we generate a total of $M$ candidate points before choosing the one with the highest EEIPU$(x)$ value. More specifically, for every stored prefix, we generate a batch of candidate points of size \texttt{b\_size} $= M/$len$(\mathbf{P})$. We deliberately include the empty prefix $[()]$, which has no fixed values (i.e. all values are randomized), to enable exploration of unknown regions.

All $M$ potential candidates are then run through our acquisition function to compute the EEIPU of each generated configuration. For every batch of generated candidate points, we use the corresponding number of memoized stages $\delta$ to only sample the expected cost of unmemoized stages from the set of GP models $\{\text{GP}_{\delta+1}, \text{GP}_{\delta+2}, \ldots, \text{GP}_K\}$. The cost of memoized stages is set to a small $\epsilon$-value, to (1) ensure numerical stability and (2) account for the overhead costs of retrieving the last-cached stage output.

The expected total cost of potential candidate $x_i$ under EEIPU is:
\begin{equation}\label{eq:mem}
    \begin{split}
        \mathbb{E}\Big[\hat{C}(x)\Big] &= \mathop{\mathbb{E}}\Big[\sum_{j=1}^K \hat{c}_j(x)\Big]
        \approx\sum_{j=1}^\delta \epsilon + \sum_{j=\delta+1}^K\mathop{\mathbb{E}}\Big[\hat{c}_j(x)\Big],
    \end{split}
\end{equation}
where the stage-wise costs $\hat{c}_j(x) \sim \mathcal{N}(\mu, K(x,x^\prime))$.
This cost discounting process serves to assign a high level of importance to cached configurations, such that, when compared to another unmemoized configuration with a relatively close expected improvement, priority would be assigned to cheaper evaluations, especially during earlier iterations of the optimization process.
\vspace{-2mm}

\subsection{Cost-Awareness with Unknown Costs}
\label{sec:cost_aware}

Our cost-aware setting is designed to incorporate memoization into our evaluation mechanism of the generated candidate observations $X^M$. \citet{snoek2012practical} and \citet{lee2020cost} briefly mention the cost modeling process for handling unknown costs, but provide no details about the inverse cost estimation nor source code to reproduce their results. Therefore we implemented EIPS and CArBO ourselves, with modifications to match our unknown-cost setting.

We define the value of EEIPU for a candidate observation $x$ as the product of its expected improvement $\text{EI}(x)$ \citep{ei} and its expected inverse cost $\mathbb{E}[\frac{1}{\hat{C}(x)}]$. We call EEIPU on every candidate observation $x_m \in X^M$ generated, passing the number of memoized stages $\delta \in [0, 1, ..., K-1]$ to account for discounted costs. The process of choosing the best candidate then goes as follows:
\begin{enumerate}
    \item We use the objective GP${\{X,Y\}}$ to compute EI($x_m$).
    \item We set the expected costs of the first $\delta$ memoized stages to $\epsilon$
    \item We apply Monte Carlo sampling to retrieve a batch of $D$ predictions from the log-cost GPs ${\{X_k, \ln(C_k)\}}\; \forall k \in \{\delta+1,2,\dots, K\}$, where $D$ is chosen to represent a fair and unbiased estimation of every stage cost.
    \item After sampling $D\times (K-\delta)$ predictions for each candidate $x_m$, we compute the total cost for each sample as $\hat{C}_d = \sum_{k=1}^{\delta} \epsilon + \sum_{k=\delta+1}^{K} \hat{c}_{d,k}\;\forall d \in \{1, 2, ..., D\}$.
    \item The total inverse cost is then estimated as:\begin{equation}\label{eq:inv_cost}
        \begin{split}
            \text{I}(x) = \mathop{\mathbb{E}}\limits_{\hat{C}(x) \sim \mathcal{N}(\mu, K(x,x^\prime))}\Big[\frac{1}{\hat{C}(x)}\Big]
            \approx \frac{1}{D}\sum_{d=1}^{D} \frac{1}{\hat{C}_d}, 
        \end{split}
    \end{equation}
\end{enumerate}

\begin{figure*}[t!]
    \centering
    \includegraphics[width=\textwidth]{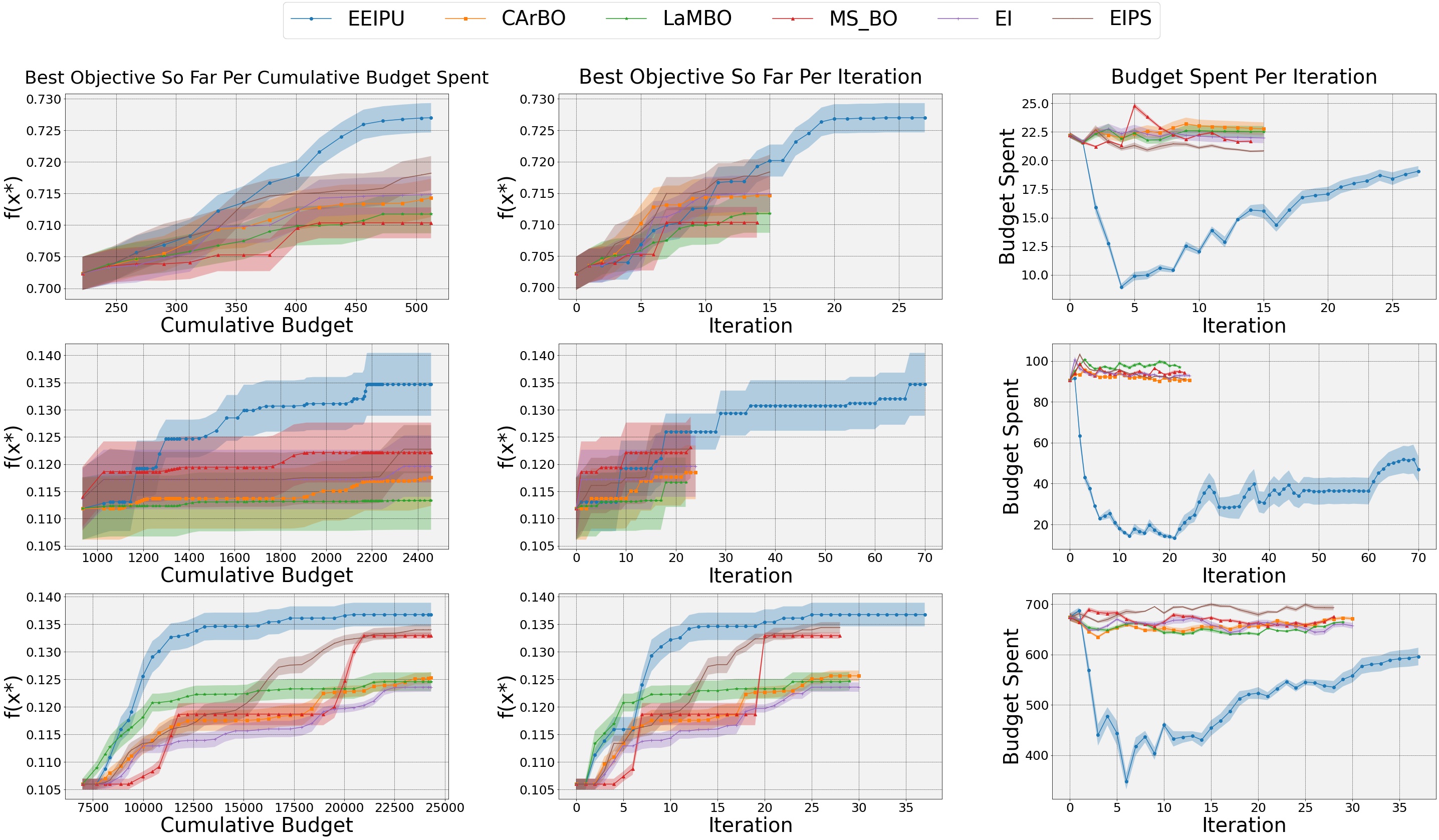}
    \caption{Real pipelines, \textbf{Top to Bottom:} Stacking, Segmentation, and T5-small Pipelines. \textbf{Left plots:} The best objective value ($f(x^*)$) achieved by each method within the cumulative consumed budget, where a quicker progression to the top means a higher objective value achieved at a relatively lower cost. \textbf{Middle plots:} The best objective value achieved with respect to the iteration count. \textbf{Right plots:} The incurred cost per iteration.}
    \label{fig:real}
\end{figure*}

\textbf{Cost-cooling}: The standard EI acquisition function has an undeniable advantage in choosing candidates with the highest expected improvement over the best objective. However, prioritizing low-cost regions throughout the optimization process may eventually lead to compromising on objective value in favor of low evaluation costs. To tackle this issue, we adopt a budget-based cost-cooling mechanism to assign a high level of importance to low-cost regions and memoized candidate observations during earlier iterations of the process, before gradually pushing EEIPU towards exploring higher-cost and entirely unexplored regions in the later iterations as the assigned budget decreases. We implement this by using an annealing schedule on the cost, adopted from the \textit{EI-cool} method \citep{lee2020cost}. Cost-cooling is defined by a factor $\eta$, which applies an exponential decay to the expected inverse cost $\text{I}(x)$. This factor is set to an initial value $\eta_0 = 1$, which is then decayed every iteration with respect to the remaining budget, such that: $\eta_t = \frac{\text{remaining budget}}{\text{total budget}}$.

 The components defined in this section are combined to propose the following formula for evaluating the Expected-Expected Improvement Per Unit of a candidate observation:
 \begin{equation}\label{eq:eeipu_cool}
     \begin{split}
         \text{EEIPU}(x) = \text{EI}(x) * \text{I}(x)^\eta, 
     \end{split}
 \end{equation}
\vspace{-8mm}
 
\subsection{EEIPU Summary}

EEIPU is illustrated in Figure~\ref{fig:eeipu}, and its pseudocode is provided in Algorithm~\ref{alg:eeipu}. EEIPU is the first multi-stage BO approach to incorporate cost- and memoization-awareness when costs are unknown beforehand. When comparing EEIPU against related methods (Table~\ref{table:ablation}), we found that none of those methods (other than EI \citep{ei}) provided source code. Thus, we manually implemented CArBO \citep{lee2020cost}, EIPS \citep{snoek2012practical},
LaMBO \cite{lin2021bayesian}, and MS\_BO \cite{kusakawa2021bayesian},
making necessary adjustments for our unknown-cost setting.
\vspace{-2mm}


\section{Main Experiments}
\label{experiments}

We demonstrate the efficiency of EEIPU on three real and three synthetic multi-stage pipelines. We report the results of our acquisition function, as well as the most closely related baselines -- EI \cite{ei}, EIPS \cite{snoek2012practical}, CArBO \cite{lee2020cost}, LaMBO \cite{lin2021bayesian} and MS\_BO \cite{kusakawa2021bayesian} -- adapted to our black-box cost, multi-stage setting to maintain a fair comparison across baselines. Because Gaussian Processes require an initial set of observations to be trained on, we ran all methods on a seeded set of $N_0 = 10$ warm-up iterations to guarantee a similar starting point for the BO process across baselines. Our experiments were repeated $T=10$ times, using different random seeds each time. Different pipeline experiments were given different optimization budgets \texttt{tot\_bgt}, set as $5 \times$ the observed average warmup budget over $T$ repeats, allowing an average of $40$ post-warmup iterations.

For every BO iteration, we generate $M = 512$ raw samples with $r = 10$ random restarts. We also sample $D = 1,000$ values from each of the cost and objective GP models. The GP prior is set to constant zero-mean, and the covariance is set to the MaternKernel, commonly used to model smooth functions. Figure
\ref{fig:real} excludes warmup iterations since they are identical for all methods.

Our experiments were conducted on a server node with 4 NVIDIA A100-SXM4-40GB GPUs and 128 CPU cores with 2 threads per core. Each method used 1 GPU and 32 CPU cores at a time, allowing us to run 4 methods concurrently. The Appendix provides more details about how we implemented EEIPU in PyTorch.

\begin{table*}[t]
\caption{Summary of real-world pipelines used in our experiments}
\label{tab:pipelines}
\vskip -0.1in
\begin{center}
\begin{small}
\begin{tabular}{lcccccr}
\toprule
Pipeline & Purpose & Number of stages & HPs per stage & Objective & Budget \\
\midrule
Stacking & 
\makecell[cl]{Classify loan applicants \\ into 'default' or 'no default'} & \makecell[cl]{2 (ensemble of 4 models, \\ logistic regression model)} & [6, 3] & AUROC & $500s$ \\
\midrule
Segmentation & \makecell[cl]{Semantic segmentation \\ of aerial images} & \makecell[cl]{3 (data preprocessing, \\ UNet model, post-processing)} & [6, 3, 3] & MIoU & $3,000s$ \\
\midrule
T5-small LM (60m) & \makecell[cl]{Fine-tuning of T5-small \\ followed by distillation} & \makecell[cl]{5 (data preprocessing, fine-tuning, \\ three distillation checkpoints)}& [1, 2, 3, 3, 3] & ROUGE score & $25,000s$ \\
\bottomrule
\end{tabular}
\end{small}
\end{center}
\end{table*}
\vspace{-2mm}

\subsection{Real ML, Vision and Language Pipelines}
\label{sec:results}

Table~\ref{tab:pipelines} describes the three AI pipelines (one ML, one vision and one language model) used in our real experiments, including the purpose of the pipeline, description of stages, dimensions, evaluation metric (objective value), and optimization budget. We measure each stage's cost as the wall-clock time to execute said stage. Each hyperparameter tuning method was run for $5$ repeats on the segmentation and T5-small pipelines, and $10$ repeats for the stacking pipeline.
We note that the AWS price of running one method on the T5-small pipeline, for the budget of $25,000$ seconds, is about \$$36$. Thus, the total AWS price of our T5-small experiment, over all $6$ methods repeated $5$ times each, is about \$$1,076$.

Figure~\ref{fig:real} shows that EEIPU (1) takes strong advantage of memoization to reduce costs, and (2) aggressively exploits lower-cost regions during early iterations, before expanding its search towards higher-cost regions. Compared to other baselines, EEIPU runs for an average of $104\%$ more iterations (Table~\ref{tab:iter}), allowing it to achieve an average of $108\%$ higher objective value (Table~\ref{tab:bestf}) within the same optimization budget.
We now discuss these results in detail, particularly how EEIPU's memoization and cost-awareness lead to quantitatively and qualitatively improved search behavior.



\begin{table}[h!]
\centering
\caption{Average number of BO iterations achieved by each method within the allocated optimization budget.}
\begin{tabular}{>{\raggedright}p{1.2cm}  >{\centering\arraybackslash}p{0.8cm}  >{\centering\arraybackslash}p{0.2cm}  >{\centering\arraybackslash}p{0.9cm}  >{\centering\arraybackslash}p{0.6cm}  >{\centering\arraybackslash}p{1cm}  >{\centering\arraybackslash}p{0.9cm}}
\toprule
Pipeline & EEIPU & EI & CArBO & EIPS & LaMBO & MS\_BO \\
\midrule
Stacking & 
$\mathbf{27}$ & $15$ & $15$ & $15$ & $15$ & $14$ \\
Segment. & 
$\mathbf{70}$ & $24$ & $24$ & $23$ & $22$ & $23$ \\
T5-small & 
$\mathbf{37}$ & $30$ & $30$ & $28$ & $29$ & $28$ \\
T5-large & 
$\mathbf{19}$ & $-$ & $12$ & $-$ & $12$ & $18$ \\
\midrule
3-synth & 
$\mathbf{166}$ & $79$ & $98$ & $75$ & $67$ & $67$ \\
5-synth & 
$\mathbf{143}$ & $72$ & $75$ & $52$ & $79$ & $94$ \\
10-synth & 
$\mathbf{110}$ & $35$ & $35$ & $36$ & $29$ & $27$ \\
\bottomrule
\end{tabular}
\label{tab:iter}
\vskip -0.2in
\end{table}

Although CArBO~\cite{lee2020cost} and EIPS~\cite{snoek2012practical} are designed as cost-aware methods, they were unable to
explore low-cost regions before exploiting high-cost ones as the budget runs out. The right-most plots in Figure~\ref{fig:real} show that only EEIPU
exhibits this behavior, which demonstrates why it is important to combine memoization with cost-awareness in the multi-stage pipeline setting. EEIPU's memoization alters the cost landscape, creating new low-cost regions (the hyperparameter ``prefixes") for exploration. Furthermore, EEIPU's sampling process explicitly generates candidates from these new low-cost regions, ensuring that they are thoroughly explored.

MS\_BO~\cite{kusakawa2021bayesian} does not perform as well as EEIPU, due to its atypical requirement that each stage outputs an objective value. In order to make MS\_BO work in our setting, we provided it with an uninformative constant objective value for all pipeline stages except the last.
Most AI pipelines (including our real ones) only produce an objective value after the final stage (e.g. AUROC or ROUGE score); intermediate stages yield no objective information.

\begin{table*}[t]
\caption{Summary of the average highest-achieved objective value within the allocated optimization budget.}
\label{tab:bestf}
\vskip -0.1in
\begin{center}
\begin{small}
\begin{tabular}{lccccccr}
\toprule
Pipeline & EEIPU & EI & CArBO & EIPS & LaMBO & MS\_BO \\
\midrule
Stacking & $\mathbf{0.727 \pm 2e^{-3}}$ & $0.715 \pm 2e^{-3}$ & $0.715 \pm 1e^{-3}$ & $0.718 \pm 3e{-3}$ & $0.712 \pm 3e^{-3}$ & $0.710 \pm 2e^{-3}$ \\
Segmentation & $\mathbf{0.135 \pm 4e^{-3}}$ & $0.120 \pm 5e^{-3}$ & $0.119 \pm 5e^{-3}$ & $0.123 \pm 4e^{-3}$ & $0.117 \pm 5e^{-3}$ & $0.123 \pm 4e^{-3}$ \\
T5-small & $\mathbf{0.137 \pm 2e^{-3}}$ & $0.124 \pm 1e^{-3}$ & $0.126 \pm 1e^{-3}$ & $0.134 \pm 2e^{-3}$ & $0.125 \pm 1e^{-3}$ & $0.133 \pm 1e^{-3}$ \\
T5-large & $\mathbf{0.159 \pm 1.1e^{-3}}$ & $-$ & $0.156 \pm 1.5e^{-3}$ & $-$ & $0.154 \pm 1.6e^{-3}$ & $0.158 \pm 1.4e^{-3}$ \\
\midrule
3-stage Synthetic
& $\mathbf{-6.90 \pm 1.39}$ & $-12.11 \pm 1.30$ & $-9.14 \pm 1.11$ & $-15.26 \pm 1.33$ & $-13.22 \pm 1.37$ & $-13.52 \pm 0.78$\\
5-stage Synthetic & $\mathbf{-2.33 \pm 0.94}$ & $-7.16 \pm 0.86$ & $-5.58 \pm 0.62$ & $-3.96 \pm 1.18$ & $-7.92 \pm 0.60$ & $-13.76 \pm 1.15$ \\
10-stage Synthetic & $\mathbf{44.20 \pm 4.63}$ & $22.51 \pm 4.57$ & $24.23 \pm 5.27$ & $27.22 \pm 4.10$ & $19.50 \pm 4.49$ & $21.96 \pm 3.90$ \\
\bottomrule
\end{tabular}
\end{small}
\end{center}
\vskip -0.1in
\end{table*}

LaMBO~\cite{lin2021bayesian} splits the hyperparameter space into disjoint regions (specifically, leaves of their so-called ``MSET tree"), and trains a separate GP for each of them, leading to exponential growth in the number of GPs with increasing hyperparameter dimension and pipeline length. However, in our budget-constrained setting, only a small number of iterations gets allocated to each LaMBO GP, resulting in low sample efficiency and thus poor hyperparameter search performance. Furthermore, although LaMBO is in principle capable of using memoization, this rarely happens because LaMBO only tries to minimize the cost of switching between sub-regions of the search space for each of the pipeline stages; however, memoization additionally requires that the exact same hyperparameter combination be regenerated for the reused stage. The issue is that LaMBO's algorithm randomly draws new hyperparameter values for each of the search space regions after triggering a switch, and at the same time, all hyperparameters in our experiments are continuous-valued. Hence, it is unlikely that the regenerated hyperparameters will exactly equal the previous hyperparameters, which prevents memoization from triggering. These reasons explain the suboptimal performance of LaMBO and MS\_BO in our benchmarks.

Overall, EEIPU outperforms other methods, yielding more iterations (Table~\ref{tab:iter}) within the allocated optimization budget, leading to state-of-the-art objective values (Table~\ref{tab:bestf}).
\vspace{-2mm}

\subsection{Larger-scale Language Model Experiments}

To show that EEIPU scales well with increasing model size in pipelines (particularly language models), we ran a subset of our experiments on the 770m-parameter T5-large language model (which is a drop-in replacement for the T5-small model in the namesake pipeline), focusing on the best-performing and most dissimilar baselines to our method --CArBO~\cite{lee2020cost}, LaMBO~\cite{lin2021bayesian}, and MS\_BO~\cite{kusakawa2021bayesian}. The experiment budget was \$1,440, costing \$72 per experiment. Each hyperparameter configuration took 51,900 seconds to run, which consumed a larger proportion of the budget compared to our smaller experiments -- hence the number of iterations achieved is lower for all methods. EEIPU ran for $19$ iterations, outlasting CArBO, LaMBO, and MS\_BO by $7, 7, \; \& \; 1$ iterations, respectively (Table~\ref{tab:iter}). EEIPU uses these additional iterations to achieve an average final objective value of $0.159$, outperforming CArBO, LaMBO, and MS\_BO, which scored $0.156, 0.154, \; \& \; 0.157$, respectively.
\vspace{-2mm}

\subsection{Longer Pipelines -- Synthetic Experiments}
In order to validate EEIPU on longer pipelines with more stages, we conducted experiments on three synthetic pipelines, whose stages are comprised of synthetic functions commonly used for evaluation in the BO literature~\cite{snoek2012practical, lee2020cost, abdolshah2019cost, multi-step-budg-bo, BAPI, lin2021bayesian, kusakawa2021bayesian}. We constructed 3-, 5-, and 10-stage pipelines (details in Appendix),
with cost functions defined per stage as a combination of \textit{cosine}, \textit{sine}, \textit{polynomial}, and \textit{logistic} functions (thus ensuring $c_k(x) \in \mathbb{R^{+}}$ across the objective function's search space). The results can be summarized as follows: compared to the baselines listed in Table~\ref{table:ablation}, EEIPU ran for an average of $149\%$ more iterations (Table~\ref{tab:iter}), and achieved $58\%$ higher objective values (Table~\ref{tab:bestf}). The results show that EEIPU performs well on longer pipelines (up to 10 stages, which is double the maximum length of our real pipelines), providing evidence that EEIPU scales to more complex AI pipelines. For space reasons, we refer readers to
the Appendix for individual performance plots of each synthetic pipeline, and other experimental details.
\vspace{-2mm}

\section{Ablation Experiments}
\label{sec:ablations}

EEIPU modifies the EI acquisition function by introducing multiple components to achieve our cost-aware, memoization-aware goals. We conduct ablation studies to show (1) the performance improvement due to these new components, and (2) to select optimal values for EEIPU parameters such as $\eta$ and $\epsilon$.
Here, we summarized the key conclusions; full details are in
the Appendix.

\paragraph{\textbf{Contributions to Objective Maximization}}
Since EEIPU targets low-cost regions first with an additional cost reduction from memoization, we wanted to ensure that it does not get stuck in low-cost areas at the expense of exploring higher-cost regions. To analyze this, we examined memoization's impact on the optimization process through experiments, detailed in Table~\ref{tab:mem_impact}.

\begin{table}[t]
\caption{Memoization improves EEIPU's performance.
}
\label{tab:mem_impact}
\vskip -0.1in
\begin{center}
\begin{small}
\begin{tabular}{lccccr}
\toprule
Pipeline & Total iter. & \makecell{Total \\ improv. iter.} & \makecell{\% Improv. \\ w/ memo.} & \makecell{\% Improv. \\ w/o memo.} \\
\midrule
Stacking & $623$ & $182$ & $27.3\%$ & $72.7\%$ \\
Segmentation & $450$ & $89$ & $24.7\%$ & $75.3\%$ \\
T5-small & $505$ & $145$ & $20.8\%$ & $79.2\%$ \\
\midrule
3-synth & $605$ & $76$ & $28.4\%$ & $71.6\%$ \\
5-synth & $474$ & $11$ & $42\%$ & $58\%$ \\
10-synth & $519$ & $122$ & $36.6\%$ & $63.4\%$ \\
\bottomrule
\end{tabular}
\end{small}
\end{center}
\vskip -0.1in
\end{table}

Table~\ref{tab:mem_impact} shows that, on average, memoization led to the best objective value in about 30\% of iterations, with non-memoized iterations achieving this in the remaining 70\%. This suggests that EEIPU maintains the crucial exploration-exploitation balance in BO, where exploration typically consumes most of the budget, and exploitation extracts the most out of previously explored regions.

\begin{figure}[t!]
    \centering
    \includegraphics[width=\linewidth]{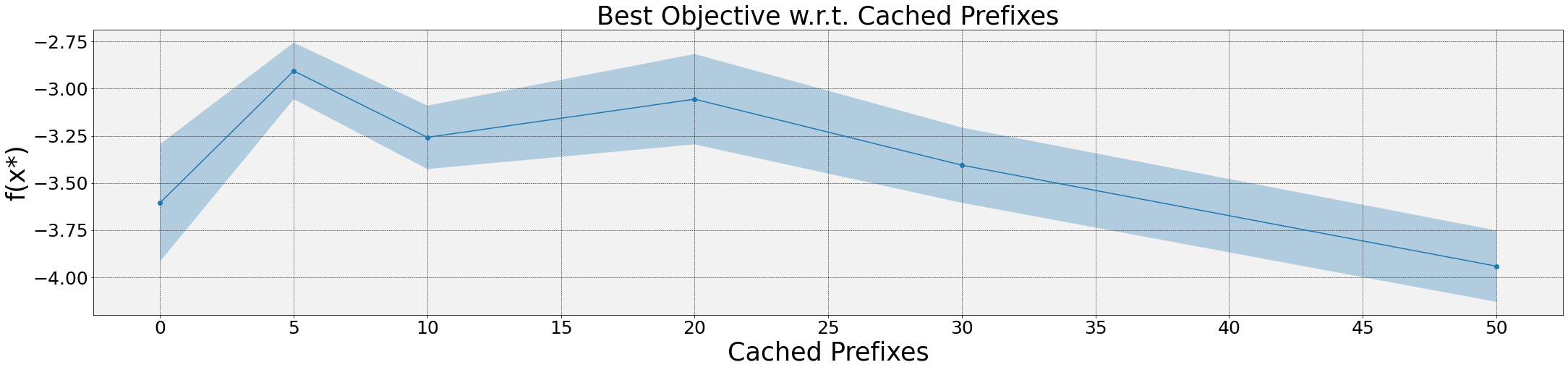}
    \vspace{-6mm}
    \caption{Best-achieved objective value w.r.t. the chosen number of cached observations.}
    \label{fig:cache}
    \vspace{-2mm}
\end{figure}

\begin{figure}[t!]
    \centering
    \includegraphics[width=\linewidth]{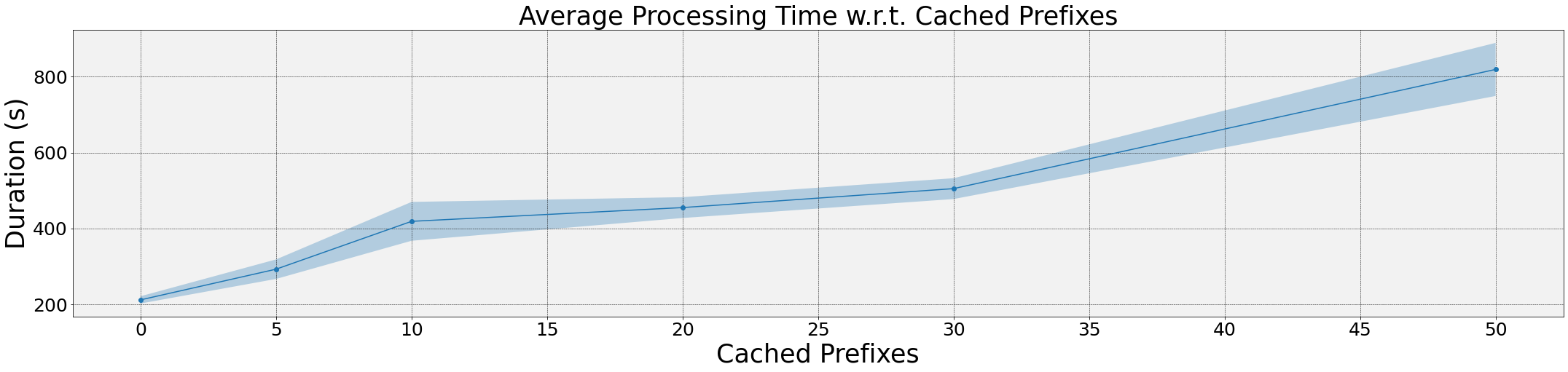}
    \vspace{-4mm}
    \caption{Duration in seconds w.r.t. the chosen number of cached observations.}
    \label{fig:dur}
    \vspace{-2mm}
\end{figure}

\paragraph{\textbf{Cache Management}}
EEIPU caches all non-complete prefixes of the top $n=5$ candidate observations, ranked by highest objective value. This choice of $n$ (1) keeps the execution time of Algorithm~\ref{alg:eeipu} low, and (2) improves EEIPU search performance because $M$, the user-specified number of search candidates to generate per BO iteration, is split across the cached prefixes -- thus, too large a value of $n$ means that some prefixes may never get searched anyway. To verify this, we ran additional experiments on our synthetic pipelines;
Figure~\ref{fig:cache} shows that using $n=5$ or 20 observations gave the best objective values, while larger values of $n$ reduced performance as we just explained. Note that one could increase $M$ in tandem with $n$ to improve performance on large values of $n$, but the increase in $M$ would also increase Algorithm~\ref{alg:eeipu}'s execution time.
Finally, we picked $n=5$ over 20 because the former takes less time: Figure~\ref{fig:dur}, which plots the time Algorithm~\ref{alg:eeipu} takes to process different $n$, shows $n=20$ takes over twice as long as $n=5$.

We also conducted an experiment (Table~\ref{tab:pref_impact}) to show that caching all possible prefixes per observation (rather than a subset), leads to better EEIPU performance. Caching all possible prefixes (denoted ``non-complete stages" in Table~\ref{tab:pref_impact}) leads to (1) more search iterations, and (2) a higher percentage of {\it iterations which improved the objective value} are because of EEIPU re-using a memoized prefix. This is because longer prefixes are more valuable to EEIPU, since they enjoy greater memoization than shorter prefixes.

\begin{table}[h!]
\caption{How the number of cached prefixes per observation impacts memoization (synthetic pipelines).}
\label{tab:pref_impact}
\vskip -0.1in
\begin{center}
\begin{small}
\begin{tabular}{lcccc}
\toprule
Memoized Stages & Total iter. & \makecell{Total \\ improv. iter.} & \makecell{\%Improv. \\ w/ memo.} & \makecell{\%Improv. \\ w/o memo.} \\
\midrule
First Stage & $455$ & $63$ & $11.1\%$ & $88.9\%$ \\
Mean Stages & $532$ & $71$ & $16.9\%$ & $83.1\%$ \\
\makecell[l]{All Non-\\ complete Stages} & $623$ & $87$ & $29.8\%$ & $70.2\%$ \\
\bottomrule
\end{tabular}
\end{small}
\end{center}
\vskip -0.1in
\end{table}

\paragraph{\textbf{Cost-cooling}}
EEIPU adjusts the cost-cooling factor $\eta_t$ based on remaining search budget, i.e. $\eta_t = \frac{\text{remaining budget}}{\text{total budget}}$. In the Appendix, we provide experiments showing that a constant cost-cooling factor $\eta_t = 1$ causes EEIPU to run more iterations but fail to transition to high-cost regions, whereas an exponential decay factor $\eta_t = 0.9 \times \eta_{t-1}$ quickly depletes the budget by rushing to high-cost areas. In contrast, EEIPU's budget-based $\eta_t$ balances exploration, achieving better optimization with fewer but more informed iterations.

\paragraph{\textbf{Memoization Overhead Costs}}
EEIPU reduces pipeline costs but incurs minor storage and retrieval overheads.
Table~\ref{tab:overhead} shows these overheads are minimal, compared to the cost of running the AI pipeline.
Even in the largest real-world experiment (T5-large), the storage and retrieval overheads are much lower than the pipeline runtime or BO process (Algorithm~\ref{alg:eeipu}) costs.

\begin{table}[h!]
\centering
\caption{Average duration (in seconds) of the pipeline, BO process, and caching overhead (real pipelines).}
\begin{tabular}{>{\raggedright}p{1.2cm}  >{\centering\arraybackslash}p{1.5cm}  >{\centering\arraybackslash}p{1.5cm}  >{\centering\arraybackslash}p{1.5cm}  >{\centering\arraybackslash}p{1.1cm}}
\toprule
Pipeline & Runtime\;\; (s) & BO Process (s) & Overhead (s) & Overhead (\%) \\
\midrule
Stacking & 
$21\pm 4.3$ & $65 \pm 5$ & $\mathbf{0.7 \pm 0.2}$ & $\sim 3.3\%$ \\
Segment. & 
$93 \pm 7.8$ & $87 \pm 6$ & $\mathbf{1.1 \pm 0.7}$ & $\sim 1.2\%$ \\
T5-base & 
$657 \pm 43$ & $146 \pm 10$ & $\mathbf{2.8 \pm 0.5}$ & $< 1\%$ \\
T5-large & 
$1,650 \pm 63$ & $181 \pm 9$ & $\mathbf{4.1 \pm 0.8}$ & $<< 1\%$ \\
\bottomrule
\end{tabular}
\label{tab:overhead}
\vskip -0.1in
\end{table}

\paragraph{\textbf{Choice of $\epsilon$}}
EEIPU computes the expected inverse cost by sampling from cost GPs. To prevent numerical issues when the discounted expected cost is very close to zero, we set memoized costs to a fixed value, $\epsilon$. We tested EEIPU's sensitivity to different $\epsilon$ values to ensure it remains unaffected by this parameter. Our results, shown in the Appendix, confirm that EEIPU is insensitive to $\epsilon$ and does not require tuning for various AI pipelines.
\vspace{-2mm}
\section{Conclusion}
\label{conclusion}

In this paper, we demonstrated that memoization, when applied to multi-stage AI pipelines, greatly improves hyperparameter search in the cost-aware setting, where the time required to run each hyperparameter configuration is just as important as the potential payoff (in terms of higher objective values).
Our resulting algorithm, EEIPU, achieves state-of-the-art results on both synthetic as well as real experiments, significantly improving the objective value from warmup iterations compared to other methods. Additionally, EEIPU incurs a substantially lower cost per iteration, achieving up to twice the BO iterations (hyperparameter evaluations) versus comparable baselines when given the same optimization budget.

\newpage



\bibliographystyle{ACM-Reference-Format}
\bibliography{main}








\newpage
\appendix

\section{APPENDIX}
\subsection{EEIPU Implementation}
\label{sec:system}


We implemented the BO algorithm part of EEIPU in \texttt{Python} using the \texttt{BoTorch} package for PyTorch. GP models are trained on normalized data, and a batch of potential candidates is generated using the \texttt{optimize\_acqf} method, which are then passed to EEIPU for the evaluation process. We begin the evaluation process by sampling expected objective and cost values from their respective GP models through MC Sampling using \texttt{botorch}'s \texttt{SobolQMCNormalSampler}, before computing the EEIPU values and returning the best-evaluated candidate configuration.

The best candidate configuration is then used to run the multi-stage pipeline, with the output of each staged cached using the \texttt{Cachestore} package on \texttt{PyPI} using the stage's configuration subset as the cache key. The output of each can be pre-processed data, an intermediate model, or an intermediate value.
To have greater control over serialization and deserialization, we utilize PyTorch's built-in \texttt{save} and \texttt{load} methods with the caching package only managing references to the files on disk.



\subsection{Additional Information for Real Pipelines}
\label{a:real-pipelines}

Table~\ref{tab:real_pipelines} provides more detailed information about the real-world pipelines in our experiments. Table~\ref{tab:pipeline_cost} shows their AWS costs.

\begin{table}[h!]
\centering
\caption{AWS costs of real-world pipelines}
\begin{tabular}{p{1.4cm} p{1.5cm} p{2.3cm} p{2cm}}
\hline
Pipeline & \# Repeats & \makecell[l]{Cost per \\ experiment in \$} & \makecell[l]{Total \\ cost in \$} \\
\hline
Stacking & $10$ & $8$ & $21.3$ \\

Segment. & $5$ & $21.5$ & $128$ \\

T5-small & $5$ & $178$ & $1066.7$ \\
\hline
\end{tabular}
\label{tab:pipeline_cost}
\end{table}

\begin{table*}[h!]
\caption{Real-world pipeline details}
\label{tab:real_pipelines}
\centering
\begin{tabular}{>{\raggedright} p{1.8cm} p{1.8cm} p{1cm} p{2.5cm} p{2.5cm} p{3cm} p{1.5cm}}
\toprule
Pipeline & Task description & Num. of stages & Stage details & Models used & Hyperparameters & Objective \\ \hline
Stacking Pipeline & Classifying loan applicants & 2 & Stage 1: Ensemble of classifiers\newline Stage 2: Logistic Regression & Extreme Gradient Boosting, Extra Trees Classifier, Random Forest, CatBoost\newline Logistic Regression & Stage 1: Combination of regularization parameter, max depth, and learning rate; across 4 models\newline Stage 2: C, tol, max iter & AUROC \\ \hline
Segmentation Pipeline & Semantic segmentation of aerial images & 3 & Stage 1: Data preprocessing\newline Stage 2: UNet Architecture\newline Stage 3: Post-Processing & Data preprocessing\newline UNet\newline Conditional Random Fields & Stage 1: means, standard deviations for the three RGB channels \newline Stage 2: batch size, learning rate, weight decay\newline Stage 3: compat, sdim, ground truth probability & Mean Intersection over Union \\ \hline
T5-small LM Pipeline (60M parameters) & Fine-tuning of T5-small followed by distillation into a smaller model & 5 & Stage 1: Data preprocessing\newline Stage 2: Fine-Tuning\newline Stages 3-5: Checkpoints of Distillation & Data preprocessing\newline Fine-Tuning (Google t5-small model)\newline Distillation Checkpoints & Stage 1: Batch size\newline Stage 2: learning rate, weight decay\newline Stages 3-5: learning rate, temperature, weight decay & ROUGE Score \\ \hline
\end{tabular}
\end{table*}

\subsection{Synthetic Pipelines}\label{sec:synthetic}

For our synthetic experiments, we defined objective and cost functions such that: 1) objective functions exhibit complex analytical forms, and 2) cost functions produce a wide range of values, for an ``uneven landscape" with regions of high and low evaluation costs.

We define the objective function $f(x)$ of a $K$-stage synthetic pipeline as the sum of $K$ synthetic BO functions, such that $f(x) = \sum_{k=1}^K f_k(x_k)$, where $f_k$ is the objective value modeling the $k^{th}$ stage of the pipeline, and $x_k$ is its corresponding set of hyperparameters. The functions $f_k$ are chosen from a compiled set of standard BO synthetic functions, defined in Table~\ref{table:objectives}. Similarly, we define one cost function per stage as a combination of \textit{cosine}, \textit{sine}, \textit{polynomial}, and \textit{logistic} functions, while ensuring $c_k(x) \in \mathbb{R^{+}}$ for every cost value across the objective function's search space.

\begin{figure*}[h!]
    \centering
    \includegraphics[width=\textwidth]{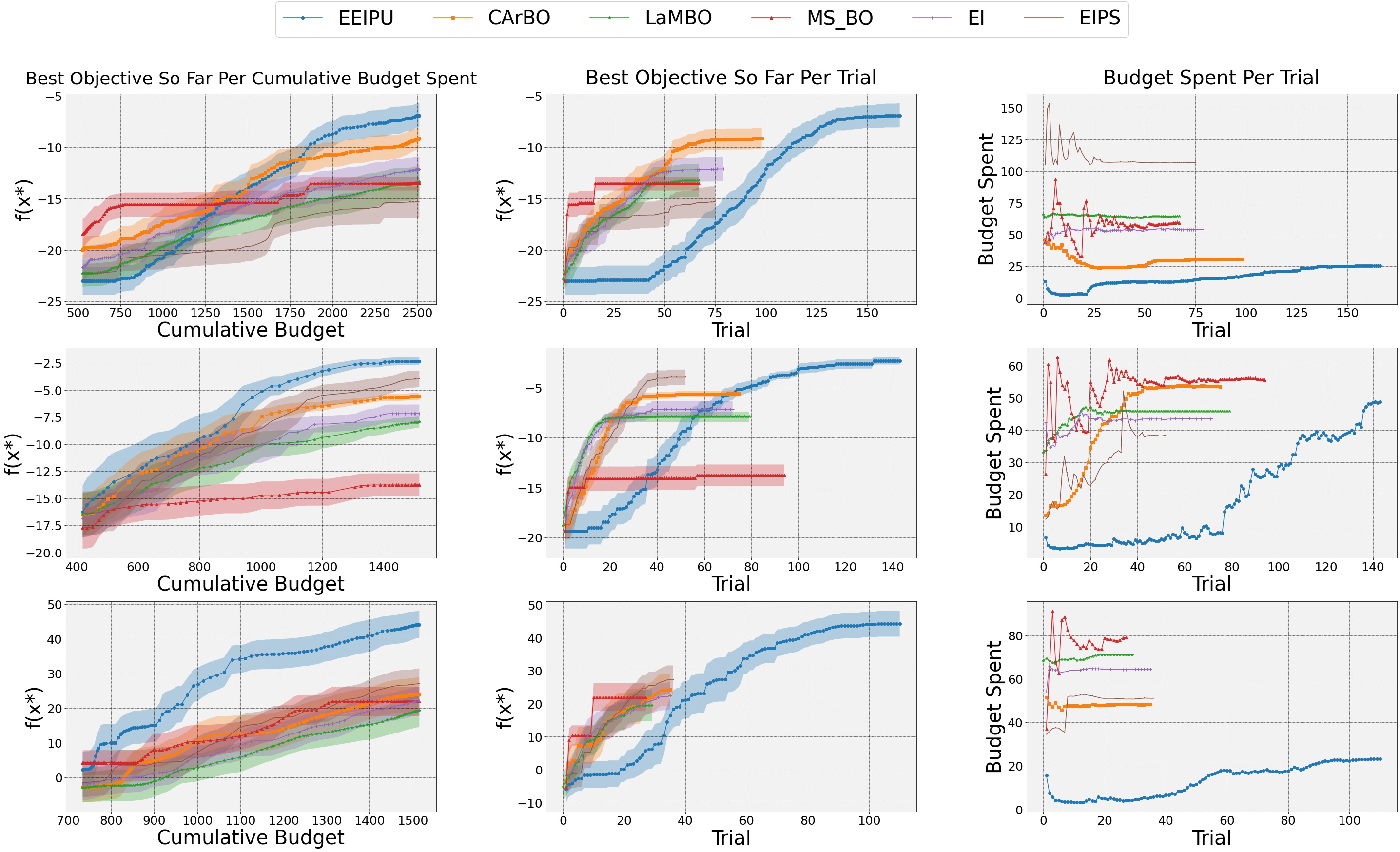}
    \caption{Synthetic pipelines, \textbf{Top to Bottom:} 3, 5, and 10-stage pipelines. \textbf{Left plots:} The best objective value ($f(x^*)$) achieved by each method within the cumulative consumed budget, where a quicker progression to the top means a higher objective value achieved at a relatively lower cost. \textbf{Middle plots:} The best objective value achieved with respect to the iteration count. \textbf{Right plots:} The incurred cost per iteration.}
    \label{fig:synthetic}
\end{figure*}

Figure~\ref{fig:synthetic} plots our synthetic experiments showing EEIPU's clear superiority over comparable methods. Within the same optimization budget, our method runs for an average of $148\%$ more iterations and improves on the best objective value from the warmup iterations by $58\%$. All results are listed in Tables \ref{tab:iter2} and \ref{tab:bestf2}.

\begin{table}[h!] \centering
\caption{Objective Functions used by synthetic pipelines}
\begin{small}
\begin{tabular}{lr}
\toprule
 Function Name & \makecell{Formula}  \\
 \midrule
 Branin-2D \citep{branin} & \scriptsize{$(x_2 - \frac{5.1}{4\pi^2}x_2^2 + \frac{5}{\pi}x_1 - 6)^2 + 10(1 - \frac{1}{8\pi})\cos(x_1) + 10$} \\
 Hartmann-3D \citep{hartmann} & \scriptsize{$- \sum_{i=1}^4 \alpha_i \exp( - \sum_{j=1}^3 A_{ij} (x_j - P_{ij})^2 )$} \\
 Beale-2D \citep{beale} & \scriptsize{$(1.5 - x_1 + x_1x_2)^2 + (2.25 - x_1 + x_1x_2^2)^2 + (2.625 - x_1 + x_1x_2^3)^2$} \\ 
 Ackley-3D \citep{ackley} & \scriptsize{$-20 \exp(-0.2 \sqrt{\frac{1}{3} \sum_{i=1}^3 x_i^2}) -
\exp(\frac{1}{3} \sum_{i=1}^3 \cos(2\pi x_i)) + 20 + e^1$} \\ 
 Michale-2D \citep{michale} & \scriptsize{$\sum_{i=1}^2 \sin(x_i) (\sin(i x_i^2 / \pi)^{20})$}\\ 
 \bottomrule
\end{tabular}
\end{small}
\label{table:objectives}
\end{table}

\begin{table}[h!]
\centering
\caption{Average number of BO iterations achieved by each method within the allocated optimization budget.}
\begin{tabular}{>{\raggedright}p{1.2cm}  >{\centering\arraybackslash}p{0.8cm}  >{\centering\arraybackslash}p{0.2cm}  >{\centering\arraybackslash}p{0.9cm}  >{\centering\arraybackslash}p{0.6cm}  >{\centering\arraybackslash}p{1cm}  >{\centering\arraybackslash}p{0.9cm}}
\toprule
Pipeline & EEIPU & EI & CArBO & EIPS & LaMBO & MS\_BO \\
\midrule
Stacking & 
$\mathbf{27}$ & $15$ & $15$ & $15$ & $15$ & $14$ \\
Segment. & 
$\mathbf{70}$ & $24$ & $24$ & $23$ & $22$ & $23$ \\
T5-small & 
$\mathbf{37}$ & $30$ & $30$ & $28$ & $29$ & $28$ \\
T5-large & 
$\mathbf{19}$ & $-$ & $12$ & $-$ & $12$ & $18$ \\
\midrule
3-synth & 
$\mathbf{166}$ & $79$ & $98$ & $75$ & $67$ & $67$ \\
5-synth & 
$\mathbf{143}$ & $72$ & $75$ & $52$ & $79$ & $94$ \\
10-synth & 
$\mathbf{110}$ & $35$ & $35$ & $36$ & $29$ & $27$ \\
\bottomrule
\end{tabular}
\label{tab:iter2}
\vskip -0.1in
\end{table}

\begin{table*}[t]
\caption{Summary of the average highest-achieved objective value within the allocated optimization budget.}
\label{tab:bestf2}
\vskip -0.1in
\begin{center}
\begin{small}
\begin{tabular}{lccccccr}
\toprule
Pipeline & EEIPU & EI & CArBO & EIPS & LaMBO & MS\_BO \\
\midrule
Stacking & $\mathbf{0.727 \pm 2e^{-3}}$ & $0.715 \pm 2e^{-3}$ & $0.715 \pm 1e^{-3}$ & $0.718 \pm 3e{-3}$ & $0.712 \pm 3e^{-3}$ & $0.710 \pm 2e^{-3}$ \\
Segmentation & $\mathbf{0.135 \pm 4e^{-3}}$ & $0.120 \pm 5e^{-3}$ & $0.119 \pm 5e^{-3}$ & $0.123 \pm 4e^{-3}$ & $0.117 \pm 5e^{-3}$ & $0.123 \pm 4e^{-3}$ \\
T5-small & $\mathbf{0.137 \pm 2e^{-3}}$ & $0.124 \pm 1e^{-3}$ & $0.126 \pm 1e^{-3}$ & $0.134 \pm 2e^{-3}$ & $0.125 \pm 1e^{-3}$ & $0.133 \pm 1e^{-3}$ \\
T5-large & $\mathbf{0.159 \pm 1.1e^{-3}}$ & $-$ & $0.156 \pm 1.5e^{-3}$ & $-$ & $0.154 \pm 1.6e^{-3}$ & $0.158 \pm 1.4e^{-3}$ \\
\midrule
3-stage Synthetic
& $\mathbf{-6.90 \pm 1.39}$ & $-12.11 \pm 1.30$ & $-9.14 \pm 1.11$ & $-15.26 \pm 1.33$ & $-13.22 \pm 1.37$ & $-13.52 \pm 0.78$\\
5-stage Synthetic & $\mathbf{-2.33 \pm 0.94}$ & $-7.16 \pm 0.86$ & $-5.58 \pm 0.62$ & $-3.96 \pm 1.18$ & $-7.92 \pm 0.60$ & $-13.76 \pm 1.15$ \\
10-stage Synthetic & $\mathbf{44.20 \pm 4.63}$ & $22.51 \pm 4.57$ & $24.23 \pm 5.27$ & $27.22 \pm 4.10$ & $19.50 \pm 4.49$ & $21.96 \pm 3.90$ \\
\bottomrule
\end{tabular}
\end{small}
\end{center}
\vskip -0.1in
\end{table*}

\subsection{Full Details of Ablation Experiments}
\label{sec:ablation}
\subsubsection{Contribution to Objective Maximization}

Because EEIPU combines an acquisition function that targets low-cost regions with a further cost discount from memoization, we want to verify that EEIPU does not exhibit pathological behavior --- such as getting stuck in low-cost parts of the landscape, and failing to explore high-cost regions. Here, we analyze the behavior of memoization and its contribution to the optimization process. We ran experiments in which we recorded statistics of memoization usage in each one of our experiments, detailed in Table~\ref{tab:memo_impact}.

Our analysis focused on computing the ratio of BO iterations that resulted in an improved objective value, distinguishing between iterations where memoization was used and those where it was not. We found that memoization was employed in a relatively low number of iterations. However, the few iterations where it was utilized led to a relatively significant percentage of objective improvements. Specifically, memoization contributes to iterations focusing on the exploitation part of the BO process by enabling EEIPU to incur a fraction of their original cost. 

Combining the set of experiments shown in Table~\ref{tab:memo_impact}, we find that memoization contributes with a newfound best objective value in an average of $\sim 30\%$ of the total number of BO iterations, with non-memoized iterations achieving a new best $\sim 70\%$ of the time. While we value the exploitation of low-cost candidate observations, our results show that EEIPU maintains the importance of maximizing the spread of exploration across the landscape, mainly leveraging memoization to improve the efficiency of the optimization process without hindering its core, while simultaneously gaining the advantage of valuable extra iterations. This behavior is integral to the BO process, where exploration occupies the majority of the budget, while exploitation steps in to extract the most out of previously explored regions of the landscape. Our results therefore show that memoization adds value to the BO process without hindering its intended behavior.

\begin{table}[h!]
\caption{Memoization improves EEIPU's performance.
}
\label{tab:memo_impact}
\vskip -0.1in
\begin{center}
\begin{small}
\begin{tabular}{lccccr}
\toprule
Pipeline & Total iter. & \makecell{Total \\ improv. iter.} & \makecell{\% Improv. \\ w/ memo.} & \makecell{\% Improv. \\ w/o memo.} \\
\midrule
Stacking & $623$ & $182$ & $27.3\%$ & $72.7\%$ \\
Segmentation & $450$ & $89$ & $24.7\%$ & $75.3\%$ \\
T5-small & $505$ & $145$ & $20.8\%$ & $79.2\%$ \\
\midrule
3-synth & $605$ & $76$ & $28.4\%$ & $71.6\%$ \\
5-synth & $474$ & $11$ & $42\%$ & $58\%$ \\
10-synth & $519$ & $122$ & $36.6\%$ & $63.4\%$ \\
\bottomrule
\end{tabular}
\end{small}
\end{center}
\vskip -0.1in
\end{table}

\subsubsection{Number of Cached Observations}
\label{sec:cache}

EEIPU works by caching the top $5$ candidate observations by associated objective values. The choice to cache a fixed number of candidates stems from the fact that the user-specified number of candidates to be generated is equally split between all stored prefixes to avoid extreme latency, meaning that the more observations are stored, the fewer candidates would be generated per cached observation. 

\begin{figure}[h!]
    \centering
    \includegraphics[width=\linewidth]{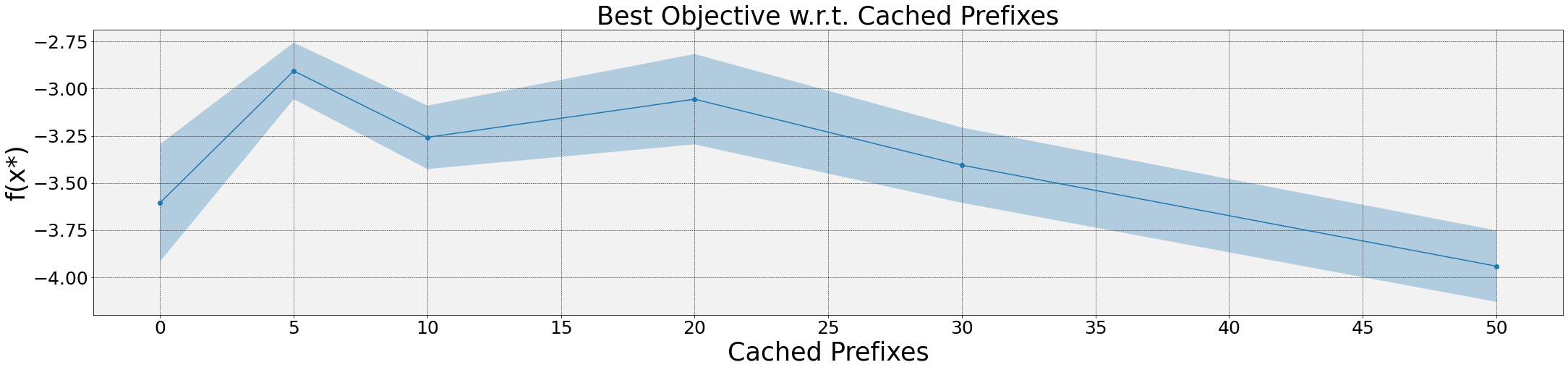}
    \vspace{-4mm}
    \caption{Best-achieved objective value w.r.t. the chosen number of cached observations.}
    \label{fig:num_cache}
    \vspace{-2mm}
\end{figure}

To support our choice of top-$5$, we conducted a series of 3 synthetic experiments where we varied the number of memoized prefixes (top-performing observations): $0, 5, 10, 20, 30, \text{and } 50$. For each experiment, we recorded the best-achieved objective value. The results are reported in Figure~\ref{fig:num_cache}:
choosing 5 or 20 memoized prefixes provided the best performance (hence, we use 5 stored prefixes in our other experiments), while larger cache sizes decreased performance significantly. This happens because EEIPU's candidate generation process works by generating random samples across all stored prefixes. The total number of random samples is fixed, so having more cached observations results in fewer generated samples per prefix, leading to worse performance. While this could be mitigated by allowing the total number of random samples to grow with cache size, this would cause the BO algorithm's computational-cost-per-iteration to increase over time (i.e. iteration number), consuming the optimization budget that would otherwise be spent on running the AI pipeline.

\begin{figure}[h!]
    \centering
    \includegraphics[width=\linewidth]{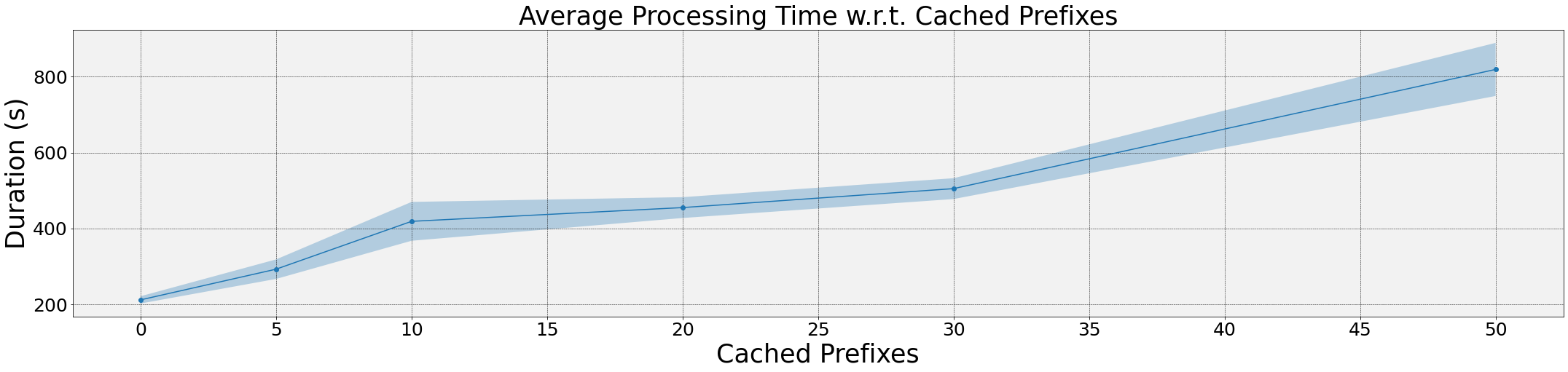}
    \vspace{-4mm}
    \caption{Duration in seconds w.r.t. the chosen number of cached observations.}
    \label{fig:dur2}
    \vspace{-2mm}
\end{figure}

As to why we chose top-5 instead of top-20 prefixes, we refer to Figure~\ref{fig:dur2}, which shows the time taken by the BO algorithm to pick a candidate, plotted against the number of cached observations. We see that the time taken increases with the number of cached observations, even though the total number of random samples remains fixed. This happens because there are other, non-sampling overheads which grow linearly with the number of cached observations.
Hence, top-5 is preferred over top-20, since the former consumes less of the optimization budget while achieving a similar objective value.
By caching 5 prefixes, we strike a balance between finding improved objective values thanks to memoization, without memoization eating too much into the optimization budget.

\subsubsection{Stage-wise Prefix Caching}

In section~\ref{sec:cache}, we showed that using the top-5 cached observations achieves the highest objective values. Here, we conduct further ablations to study how the number of pipeline stages cached per observation affects EEIPU's performance. Normally, EEIPU's algorithm stores every non-complete prefix, i.e. stages $1$ to $K-1$ in a $K$-stage pipeline, allowing it to retrace any number of stages before taking a new direction. One might ask if it is potentially better to store a limited number of prefixes per observation (instead of all prefixes), so that more observations might be cached.

To test this, we varied the number of stages stored per EEIPU observation, in the following configurations:
\begin{itemize}
    \item $|\textbf{P}| = K-1$, caching all stages except the last. This is the ``default" EEIPU behavior.
    \item $|\textbf{P}| = 1$, caching the first stage alone.
    \item $|\textbf{P}| = \ceil*{\frac{K}{2}}$, caching the mean number of stages.
\end{itemize}
Using these configurations, we ran several trials of our synthetic pipelines. Table~\ref{tab:pref_imp} summarizes the results; our default approach (store all stages except the last) significantly outperforms the other methods in two respects: (1) it achieves more total BO iterations (i.e. each iteration costs less on average because there are more memoized stages to re-start from), and (2) a greater percentage of those iterations improved upon the best objective value achieved so far (i.e. having more memoized stages improves the effectiveness of the hyperparameter search).

The results suggest that increasing the number of cached prefix stages also increases the probability that EEIPU will select a later stage (i.e. longer prefix) when starting a new iteration. Intuitively, the more stages are available, the better the odds of finding high-objective branches in some subspaces (this is essentially what prefixes are) of the full, high-dimensional hyperparameter search space.
\begin{table}[h!]
\caption{How the number of cached prefixes per observation impacts memoization (synthetic pipelines).}
\label{tab:pref_imp}
\vskip -0.1in
\begin{center}
\begin{small}
\begin{tabular}{lcccc}
\toprule
Memoized Stages & Total iter. & \makecell{Total \\ improv. iter.} & \makecell{\%Improv. \\ w/ memo.} & \makecell{\%Improv. \\ w/o memo.} \\
\midrule
First Stage & $455$ & $63$ & $11.1\%$ & $88.9\%$ \\
Mean Stages & $532$ & $71$ & $16.9\%$ & $83.1\%$ \\
\makecell[l]{All Non-\\ complete Stages} & $623$ & $87$ & $29.8\%$ & $70.2\%$ \\
\bottomrule
\end{tabular}
\end{small}
\end{center}
\vskip -0.1in
\end{table}

\subsubsection{Cost-cooling}

\begin{figure*}[h!]
    \centering
    \includegraphics[width=\textwidth]{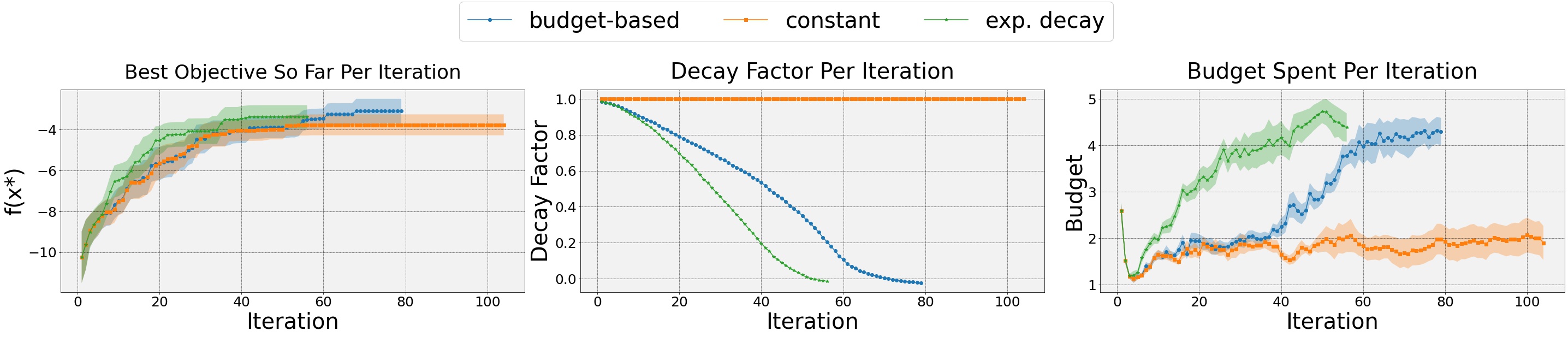}
    \caption{Decay Factor Ablation Experiments. EEIPU was run on a set of different decay formulas, ranging from the budget-based formula $\eta = \frac{\texttt{remaining budget}}{\texttt{total budget}}$, as well as an exponential decay formula $\mathbf{\eta_{t} = 0.9\;.\;\eta_{t-1}}$, and a constant factor $\mathbf{\eta = 1}$. \textbf{(1)} The left plot shows the best objective achieved per iteration, \textbf{(2)} The middle plot shows the changes of $\eta$ with respect to BO iterations, \textbf{(3)} The right plot shows the changes of accessed cost regions per iteration.}
    \label{fig:eta}
\end{figure*}

One important aspect of EEIPU, inspired by the CArBO method~\cite{lee2020cost}, is the cost-cooling factor $\eta$. The purpose of this variable, explained in Section 3.2 of the main paper, is to prioritize low-cost regions during initial BO iterations, gradually decaying the cost exponent every iteration $t$ to encourage the exploration of high-cost regions towards the end of the optimization process. To guarantee this behavior, $\eta$ varies with respect to the allocated budget such that $\eta_t = \frac{\text{remaining budget}}{\text{total budget}}$.

In this section, we test EEIPU's sensitivity to the choice of $\eta$. We ran ablation experiments using a constant factor and an exponent factor chosen from separate experiments on values $\{0.99, 0.95, 9.9, 0.7\}$. The chosen values are the following:
\begin{itemize}
    \item $\eta_t = \frac{\text{remaining budget}}{\text{total budget}}$
    \item $\eta_t = 0.9 \;\times\; \eta_{t-1}$ 
    \item $\eta_t = 1$
\end{itemize}

Figure~\ref{fig:eta} shows that the constant factor $\eta_t = 1$ runs for the highest number of iterations. However, it rarely ever seems to access any region with a cost higher than 2 units. This behavior is counter-intuitive and opposing to the intention behind BO processes, which causes the noticed stagnation towards the end of the optimization process. The exponential decay factor $\eta = 0.9 \times \eta_{t-1}$ displays the opposite behavior. It prematurely rushes to the high-cost regions, making very limited use of memoization and quickly exhausting the allocated budget without finding the optimizer. The budget-based factor, on the other hand, displays steady decline compared to the exponential decay factor, and allows half the process for low-cost regions before gradually shifting towards high-cost areas. While it achieves less iterations than the naive constant factor, it does so with smarter budget management and more acquired information of the search space, allowing the extra iterations ---compared to the exponential decay--- to lead to the best-achieved objective value.

Decaying $\eta$ with respect to the budget guarantees the expected steady behavior, decaying almost linearly before dropping at a higher pace towards later iterations. Therefore, we believe this technique to be robust enough to perform in theory as well as practice.

\subsubsection{Memoization Overhead Costs}

While EEIPU is able to discount the cost of running the full pipeline, it also incurs cache storage and retrieval overheads, which consume a small part of the optimization budget. Here, we report statistics collected from our real experiments (Table~\ref{tab:mem_overhead}), to demonstrate that EEIPU's storage and retrieval overheads are small relative to the AI pipeline.

The columns marked as "Overhead" are defined as storage plus retrieval costs. The percentage calculation is defined as $\frac{\text{Overhead}}{\text{Runtime}}$; here, Runtime refers to running the AI pipeline, and does not include the cost of the BO Process, which we report in a separate column. The results indicate that, even for our largest real-world experiment (T5-Large), EEIPU's storage and retrieval costs are much smaller than either the pipeline Runtime or the cost of running the BO Process (BO algorithm).

\begin{table}[h!]
\centering
\caption{Average duration (in seconds) of the pipeline, BO process, and caching overhead (real pipelines).}
\begin{tabular}{>{\raggedright}p{1.2cm}  >{\centering\arraybackslash}p{1.5cm}  >{\centering\arraybackslash}p{1.5cm}  >{\centering\arraybackslash}p{1.5cm}  >{\centering\arraybackslash}p{1.1cm}}
\toprule
Pipeline & Runtime\;\; (s) & BO Process (s) & Overhead (s) & Overhead (\%) \\
\midrule
Stacking & 
$21\pm 4.3$ & $65 \pm 5$ & $\mathbf{0.7 \pm 0.2}$ & $\sim 3.3\%$ \\
Segment. & 
$93 \pm 7.8$ & $87 \pm 6$ & $\mathbf{1.1 \pm 0.7}$ & $\sim 1.2\%$ \\
T5-base & 
$657 \pm 43$ & $146 \pm 10$ & $\mathbf{2.8 \pm 0.5}$ & $< 1\%$ \\
T5-large & 
$1,650 \pm 63$ & $181 \pm 9$ & $\mathbf{4.1 \pm 0.8}$ & $<< 1\%$ \\
\bottomrule
\end{tabular}
\label{tab:mem_overhead}
\vskip -0.1in
\end{table}

\subsubsection{Choice of $\epsilon$}

During the BO process, the expected inverse cost is computed by sampling from cost-GPs, before estimating the expected inverse cost. When discounting memoized costs, we set their respective expectations to a fixed value $\epsilon$; this is a precautionary measure to avoid numerical issues that randomly occur if the expected inverse cost is too close to $0$. Ideally, we want to show that EEIPU is insensitive to different choices of $\epsilon$. Thus, we ran experiments on different values of $\epsilon$ to analyze EEIPU's sensitivity, if any, to it.

\begin{figure}[h!]
    \centering
    \includegraphics[width=\linewidth]{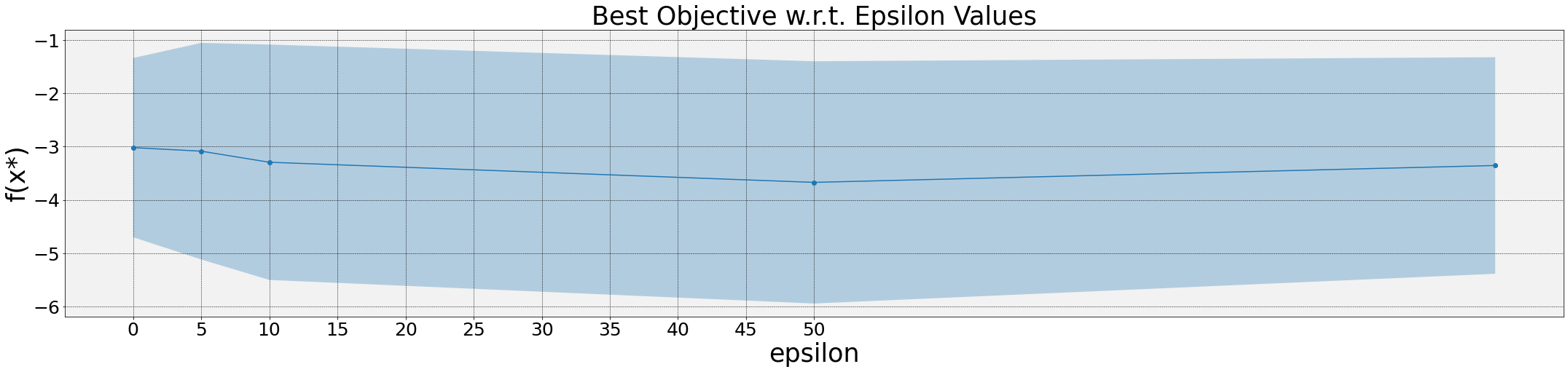}
    \caption{Best objective achieved with different $\epsilon$ values.}
    \label{fig:eps_abl}
\end{figure}

We set $\epsilon$ to a range of different values, ranging from $0$ to $100$. Our results, shown in Figure~\ref{fig:eps_abl}, demonstrate EEIPU's insensitivity to the chosen $\epsilon$ value -- in other words, it does not need to be tuned for different AI pipelines.

\subsection{Limitations}
\label{a:limitations}

When the objective function we are optimizing for is positively correlated with the cost function, the optimal objective value $f(x^*)$ would be located at a very high-cost region. In this case, sample-efficient techniques, such as EI \citep{ei} could potentially find the optimal objective value by immediately targeting the high-cost regions where the expected improvement is maximized, while EEIPU is initially directed towards low-cost regions for a large part of the optimization process. Certain measures could be taken to overcome this limitation, such as implementing a warmup phase to prioritize hyperparameter choices in high-variance/uncertain regions of the search space. Intuitively, this variation of EEIPU initially prioritizes the acquisition of cost and objective information over a wide search space. When the objective is correlated with cost, this allows high-cost-high-objective regions to be quickly discovered in earlier iterations, without wasting iterations on low-cost-low-objective regions.




\end{document}